\documentclass[acmtog]{acmart}

\renewcommand\footnotetextcopyrightpermission[1]{} % removes footnote with conference information in first column

\usepackage{xspace}
\usepackage{booktabs} % For formal tables
\usepackage{enumitem}
\usepackage{makecell} % 引入 makecell 包
\newcommand{\ModelName}{\emph{Cobra}\xspace}

\usepackage{multirow}
\AtBeginDocument{%
  }

\acmJournal{TOG}
\begin{document}

%%
%% The "title" command has an optional parameter,
%% allowing the author to define a "short title" to be used in page headers.
\title{Cobra: Efficient Line Art COlorization with BRoAder References}

%%
%% The "author" command and its associated commands are used to define
%% the authors and their affiliations.
%% Of note is the shared affiliation of the first two authors, and the
%% "authornote" and "authornotemark" commands
%% used to denote shared contribution to the research.
\author{JUNHAO ZHUANG}
\orcid{0000-0002-1642-0750}
% \author{G.K.M. Tobin}
% \authornotemark[1]
% \email{webmaster@marysville-ohio.com}
\affiliation{%
  \institution{Tsinghua University}
  % \city{Shenzhen}
  \country{China}
}
\email{zhuangjh23@mails.tsinghua.edu.cn}

\author{LINGEN LI}
\orcid{0000-0002-1313-8717}
\affiliation{%
  \institution{The Chinese University of Hong Kong}
  % \city{Hong Kong}
  \country{China}
}
\email{lgli@link.cuhk.edu.hk}
\author{XUAN JU}
\orcid{0000-0002-0668-1375}
\affiliation{%
  \institution{The Chinese University of Hong Kong}
  % \city{Hong Kong}
  \country{China}
}
\email{juxuan.27@gmail.com}
\author{ZHAOYANG ZHANG}
\orcid{0009-0003-5583-6454}
\affiliation{%
  \institution{Tencent ARC Lab}
  % \city{Shenzhen}
  \country{China}
}
\authornote{Project Lead.}
\email{zhaoyangzhang@link.cuhk.edu.hk}
\author{CHUN YUAN}
\orcid{0000-0002-3590-6676}
\affiliation{%
  \institution{Tsinghua University}
  % \city{Shenzhen}
  \country{China}
}
\authornote{Co-corresponding authors: Chun Yuan and Ying Shan.}
% \authornotemark[2]
\email{yuanc@sz.tsinghua.edu.cn}
\author{YING SHAN}
\orcid{0000-0001-7673-8325}
\affiliation{%
  \institution{Tencent ARC Lab}
  % \city{Shenzhen}
  \country{China}
}
\authornotemark[2]
\email{yingsshan@tencent.com}
%%
%% By default, the full list of authors will be used in the page
%% headers. Often, this list is too long, and will overlap
%% other information printed in the page headers. This command allows
%% the author to define a more concise list
%% of authors' names for this purpose.
% \renewcommand{\shortauthors}{Trovato et al.}

%%
%% The abstract is a short summary of the work to be presented in the
%% article.
\begin{abstract}
  % Reference-based image colorization is a critical task in the animation production industry, demanding high accuracy, efficiency, supported context length, and usage flexibility. 
  % While diffusion models have advanced image generation capability,their use in reference-based colorization remains limited and insufficient for industrial application.
  % Existing methods often lack the capability to handle a large number of reference images, require time-consuming inference time, and do not support diverse input formats.
  % In this paper, we introduce \ModelName, an efficient, precise, and versatile image colorization method that supports up to 200 reference images and color hints as input with low latency.
  % \ModelNameuses a specialized causal attention mechanism and positional encoding to manage long reference contexts efficiently while ensuring identity consistency through in-context learning ability in attention mechanism.
  % \todo{add more details and insight for cobra}
  % Comprehensive results show \ModelNameachieves broader context referencing and faster colorization while achieving precise reference-based image colorization, addressing key industrial requirements for reference-based colorization.
The comic production industry requires reference-based line art colorization with high accuracy, efficiency, contextual consistency, and flexible control.
A comic page often involves diverse characters, objects, and backgrounds, which complicates the coloring process.
Despite advancements in diffusion models for image generation, their application in line art colorization remains limited, facing challenges related to handling extensive reference images, time-consuming inference, and flexible control.
We investigate the necessity of extensive contextual image guidance on the quality of line art colorization. To address these challenges, we introduce \ModelName, an efficient and versatile method that supports color hints and utilizes over 200 reference images while maintaining low latency. 
Central to \ModelName is a Causal Sparse DiT architecture, which leverages specially designed positional encodings, causal sparse attention, and Key-Value Cache to effectively manage long-context references and ensure color identity consistency.
Results demonstrate that \ModelName achieves accurate line art colorization through extensive contextual reference, significantly enhancing inference speed and interactivity, thereby meeting critical industrial demands. 
We release our codes and models on our project
page: \url{https://zhuang2002.github.io/Cobra/}.

\end{abstract}

%%
%% The code below is generated by the tool at http://dl.acm.org/ccs.cfm.
%% Please copy and paste the code instead of the example below.
%%


%%
%% Keywords. The author(s) should pick words that accurately describe
%% the work being presented. Separate the keywords with commas.

% \received{20 February 2007}
% \received[revised]{12 March 2009}
% \received[accepted]{5 June 2009}

%%
%% This command processes the author and affiliation and title
%% information and builds the first part of the formatted document.

\begin{teaserfigure}
% \vspace{-1mm}
\includegraphics[width=1.0\textwidth]{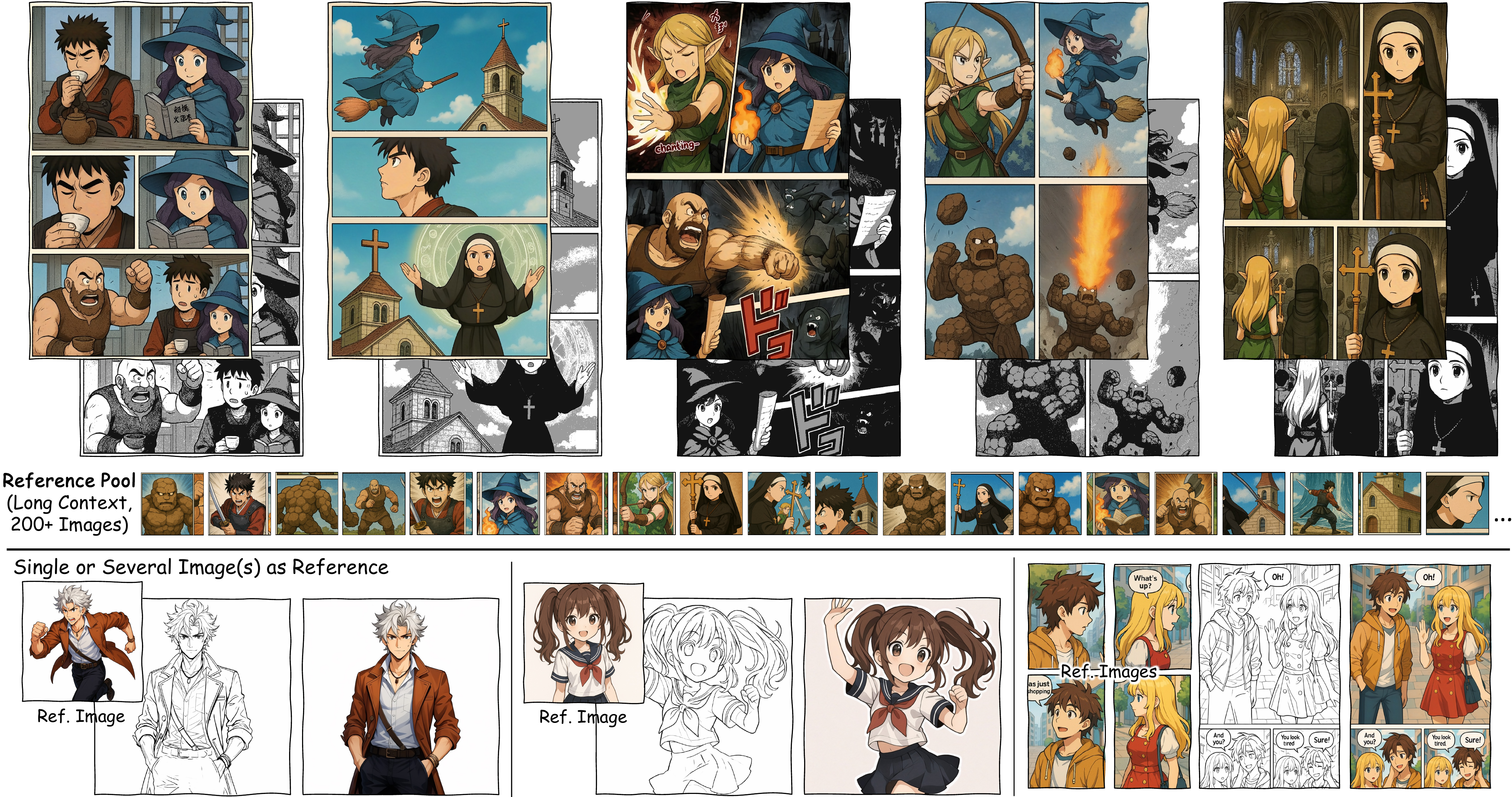}
% \vspace{-10mm}
\caption{
\textbf{\ModelName} is a novel efficient long-context fine-grained ID preservation framework for line art colorization, achieving high precision, efficiency, and flexible usability for comic colorization. By effectively integrating extensive contextual references, it transforms black-and-white line art into vibrant illustrations. }
% The comic images presented in this paper are generated by AI.}
\end{teaserfigure}

\maketitle
\section{Introduction}\label{sec:introduction}
\textcolor{black}{Diffusion models~\cite{rombach2022high, zhang2023adding} have transformed image generation, inpainting~\cite{zhuang2023task, ju2024brushnet}, and editing~\cite{brooks2023instructpix2pix, hertz2022prompt}, yet their application in multi-reference-based colorization, especially for industrial-scale tasks, remains underexplored.}

\textcolor{black}{Early solutions for line art colorization focused on palettes, color hints, and text control mechanisms. Palettes~\cite{chang2015palette, wang2022palgan} provide consistency but limit flexibility in diverse comic styles. Color hint methods~\cite{xian2018texturegan, yun2023icolorit, liang2024control, zhang2018two}
offer adaptability, yet they lack the automation required for rapid industrial applications. Text control~\cite{zhang2023adding, bahng2018coloring, chang2022coder} enables intuitive guidance but the text encoder is computationally expensive and sensitive to input clarity.}

\textcolor{black}{Existing reference-based colorization methods~\cite{cao2023animediffusion, zhuang2024colorflow, comicolorization} address these challenges but are limited by the number of reference images, slow inference times, and an inability to handle complex, context-dependent comic pages. ScreenVAE~\cite{xie2020manga} extracts style vectors from a single reference image for manga colorization, while ColorFlow~\cite{zhuang2024colorflow} employs a three-stage framework with dual-branch networks, limited to 12 references. AniDoc~\cite{meng2024anidoc} focuses on single-character anime video colorization.}
However, comic pages feature diverse items, backgrounds, and character details that are context-dependent and dispersed across various pages. Ensuring color ID consistency requires long-context references. Current methods are inadequate for practical industrial applications due to their limited reference images, lengthy inference times, and lack of support for diverse input formats, such as color hint references.

To address these challenges, we introduce \ModelName, a novel long-context line art colorization framework that offers high precision, efficiency, and flexible usability. It supports over 200 reference images and color hints as input. Rich contextual references are crucial for comic art colorization, as shown in Tab. \ref{tab:ref_num}. \ModelName enhances accuracy and efficiency through four key innovations:
(1) \textbf{Multi-Identity Consistency}: By spatially concatenating clean reference image latents with noise latents, we use attention mechanisms to maintain consistency between reference images and target outputs.
(2) \textbf{Efficient Attention Design}: 
\textcolor{black}{We sparsify traditional full attention by eliminating pairwise attention computations among reference images, thereby reducing redundant interactions. Inspired by advances in language modeling~\cite{vaswani2023attentionneed, yenduri2023GPT1, pope2022efficientlyscalingtransformerinference}, we further adopt a causal attention mechanism along with a KV-cache, which significantly reduces memory and computational costs. These design choices collectively improve inference efficiency without sacrificing colorization quality.}
(3) \textbf{Flexible Position Encoding}: 
The Localized Reusable Position Encoding allows reuse of local positional encodings without altering pre-trained DiT 2D encodings, accommodating an arbitrary number of references within a constrained resolution.
%
% Moreover, \ModelName~ supports the integration of color hints, ensuring flexibility of use.
(4) \textbf{Color Hint Integration}: \ModelName also supports the integration of color hints, ensuring flexible usage and enhancing the line
 art colorization process.

To comprehensively evaluate comic page line art colorization, we introduce \textit{Cobra-bench}, which includes 30 comic chapters, each with 50 line art images for colorization and 100 reference images. 
Extensive evaluations demonstrate that \ModelName~surpasses existing baselines in image quality, color ID accuracy, and inference efficiency, particularly with richer contextual information. Cobra also requires minimal inference modifications to adapt to anime video colorization. Our contributions are summarized as follows:

\begin{itemize}[itemsep=0pt,leftmargin=*]
\item We introduce Cobra, an innovative framework designed for long-context comic line art colorization, achieving high precision, efficiency, and flexible control, supporting both color hints and long reference contexts. 

\item We present the Causal Sparse DiT architecture, featuring Localized Reusable Position Encoding and Causal Sparse Attention with KV-Cache. This design significantly reduces the computational complexity of traditional full attention mechanisms, enabling the processing of over 200 reference images with lower overhead and reduced latency. 

\item We establish \textit{Cobra-bench}, a comprehensive benchmark for multi-reference-based comic line art colorization. Extensive evaluations show that broader image references are crucial for high-quality comic colorization, with our method significantly outperforming existing techniques in image quality and color ID accuracy.
\end{itemize} 

\section{Related Work}\label{sec:related_work}

\subsection{Image Colorization}

Image colorization \cite{zhang2018two,comicolorization} converts grayscale images, such as manga \cite{qu2006manga}, line art \cite{kim2019tag2pix}, sketches \cite{zhang2023adding}, and natural grayscale images \cite{zabari2023diffusing}, into colored images. Previous methods rely on text \cite{zabari2023diffusing,zhang2023adding,weng2024cad,chang2023coins} and palette controls \cite{utintu2024sketchdeco,wu2023flexicon,wang2022palgan,xiao2020example} or require manual color hints (e.g., scribbles) \cite{dou2021dual,zhang2021user,zhang2018two,yun2023icolorit,ci2018user,liu2018auto}, but these approaches only provide rough color styles and fail to ensure accurate color preservation for identities in black-and-white images. To improve controllability, reference-based colorization \cite{cao2023animediffusion,zou2024lightweight,wu2023flexicon,wang2023unsupervised,wu2023self,comicolorization}, which uses reference images for guidance, has become popular. For instance, ColorFlow \cite{zhuang2024colorflow} uses a three-stage framework for retrieval, colorization, and upscaling, while AniDoc \cite{meng2024anidoc} employs a dual-branch 3D UNet framework with an animation character image as a reference for line art video colorization.

Despite these advancements, existing methods face limitations in practical applications due to the diverse objects and characters on a single comic page and challenges in retrieval accuracy. They often struggle with limited reference images, long inference times, and inadequate support for varied input formats like color hint references. 
To overcome these issues, we propose Cobra, which uses Localized Reusable Position Encoding and Causal Sparse Attention to enable efficient long-context reference, resulting in more accurate and faster comic line art colorization.

% Despite these advancements, existing methods exhibit limitations in practical industrial applications, primarily due to the rich diversity of objects and character content within a single comic page, compounded by challenges in retrieval accuracy. Current approaches often struggle with a limited number of reference images, lengthy inference times, and insufficient support for varied input formats, such as color hint references.
% To address these challenges, we propose Cobra, which integrates Localized Reusable Position Encoding and Causal Sparse Attention to facilitate efficient long-context reference, thereby enabling more accurate and rapid comic line art colorization.

\begin{figure*}[t]
    \centering
    % \vspace{-3mm}
    \includegraphics[width=1.0\textwidth]{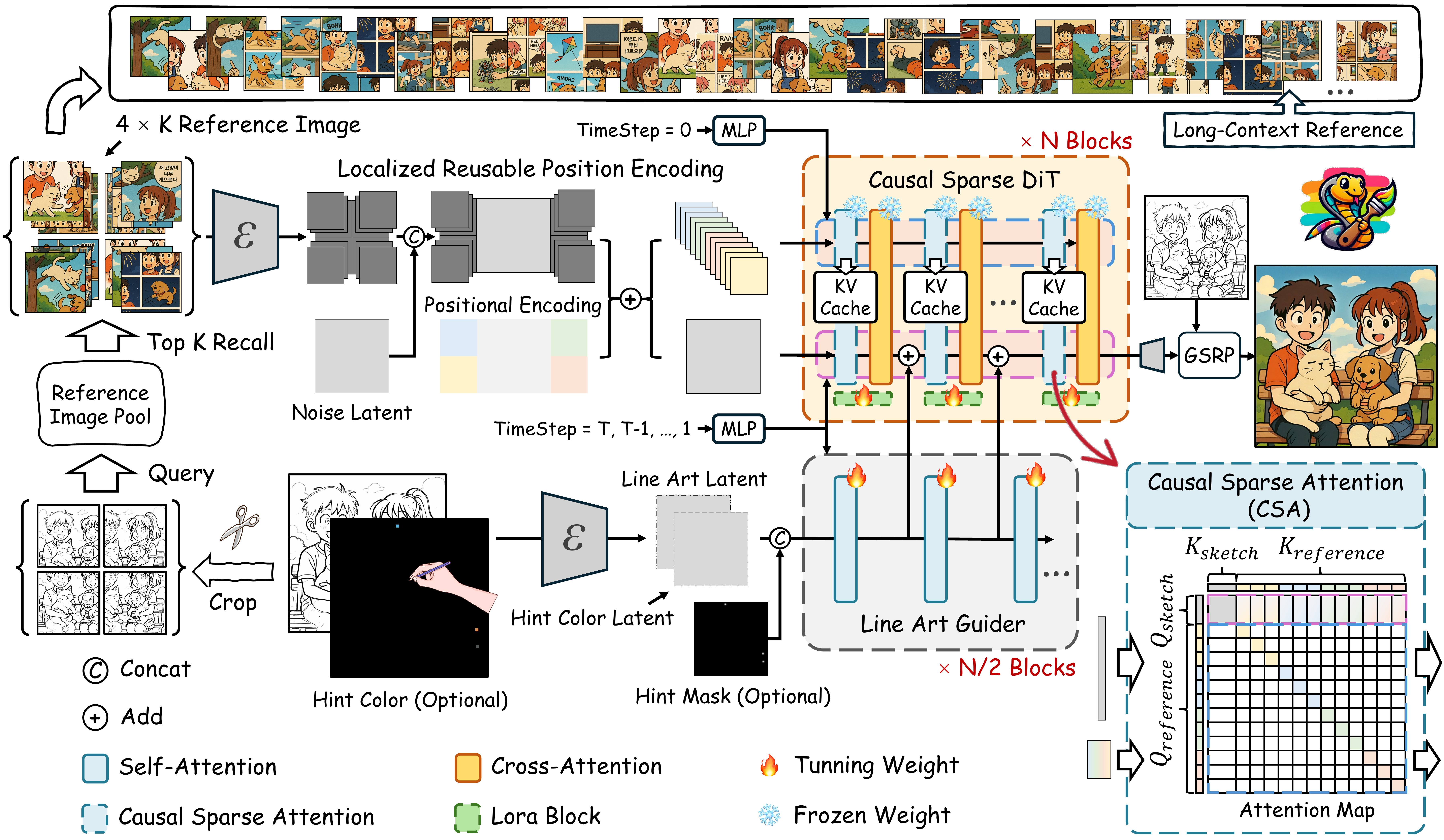}
    % \vspace{-2mm}
    \caption{
    \textbf{The overview of \ModelName.}
This figure depicts the framework of \ModelName, which utilizes a large collection of retrieved reference images to guide the colorization of comic line art. The framework effectively manages an arbitrary number of contextual image references through localized reusable positional encoding, ensuring appropriate aspect ratios and resolutions. Additionally, the causal sparse DiT architecture processes long contextual references, enhancing identity preservation and color accuracy while reducing computational complexity. The integration of optional color hints further ensures user flexibility, culminating in high-quality coloring that is highly suitable for industrial applications.
    }
    % \vspace{-2mm}
    \label{fig:FlowChat}
\end{figure*}

\subsection{ldentity Preservation} 
% ID-Preservation is a trending topic in the field of generative models. Previous approaches can be primarily divided into two categories: one involves fine-tuning the generative model to remember one or more predefined concepts~\cite{1,2,3}, while the other utilizes plug-and-play modules trained on large-scale datasets to control the generation of desired concepts using given image content during inference. Overall, prior methods typically focus on a limited set of predefined concepts to maintain. In contrast, our proposed ColorFlow offers a robust and automated three-stage framework for sequential image colorization, effectively handling the rich and varied characters, objects, and backgrounds found in comic sequences, making it suitable for industrial applications.
Identity preservation enables users to provide reference images to generate results containing the given identities. Traditional approaches can be classified into tuning-based and plug-and-play methods. 
The tuning-based methods~\cite{ruiz2023dreambooth,gal2022image,jiang2024videobooth} are designed to optimize the weight of the diffusion model to memorize the identity information, which is limited by the consuming online fine-tuning time, large efforts for manually collecting training samples, and the poor generalization ability.
Plug-and-play approaches~\cite{chen2024anydoor,gal2023encoder,ye2023ip,li2024photomaker,wang2024instantid} usually train an encoder to map reference images to visual representations and use them to guide the diffusion generation process. 
Although these methods are relatively easy to use, numerous studies~\cite{ju2023humansd,deltatuning} have demonstrated that the adapter-based approaches they rely on often lead to suboptimal performance.
Recent research on In-Context LoRA~\cite{huang2024context} has demonstrated that the in-context learning capabilities of self-attention can effectively and naturally preserve identities, resulting in superior performance.

However, the resolution limitations of 2D positional encodings in pre-trained diffusion model  and the quadratic complexity of full attention restrict In-Context LoRA to generating a limited number of related images. Our approach overcomes these limitations by reusing localized positional encoding patches to handle any number of references and introducing Causal Sparse Attention with KV-Cache for efficient linear complexity relative to the number of reference images.

% , combined with the computational complexity that increases quadratically with the number of images in full attention, restrict In-Context LoRA to generating only a limited number of interrelated images. In contrast, our approach utilizes the reuse of localized positional encoding patches to accommodate an arbitrary number of references. Additionally, we introduce Causal Sparse Attention in conjunction with KV-Cache, achieving efficient linear complexity in relation to the number of reference images.

% Thus, in \ModelName, we directly concatenate reference images along the spatial dimension and leverage the strong in-context learning ability of attention to index identities.
\section{Method}\label{sec:method}
We present \ModelName, an efficient long-context fine-grained ID preservation framework designed for line art colorization, providing a robust solution suitable for industrial applications. As illustrated in Fig. \ref{fig:FlowChat}, \ModelName\ retrieves relevant images from a reference image pool, denoted as \( R = \{ r_1, r_2, \ldots, r_n \} \). Each reference image \( r_i \) contributes to the colorization of a given line art image \( L \), guiding the coloring process to enhance the final output. Users can also incorporate color hints to further refine the results. The architecture of \ModelName\ consists of three core components: Causal Sparse DiT, Localized Reusable Position Encoding, and Line Art Guider, as shown in Fig. \ref{fig:FlowChat}.
\subsection{Causal Sparse DiT}
% \subsubsection{Causal Sparse Attention}
\begin{figure}[t]
    % \centering
    % \vspace{-3mm}
    \includegraphics[width=0.48\textwidth]{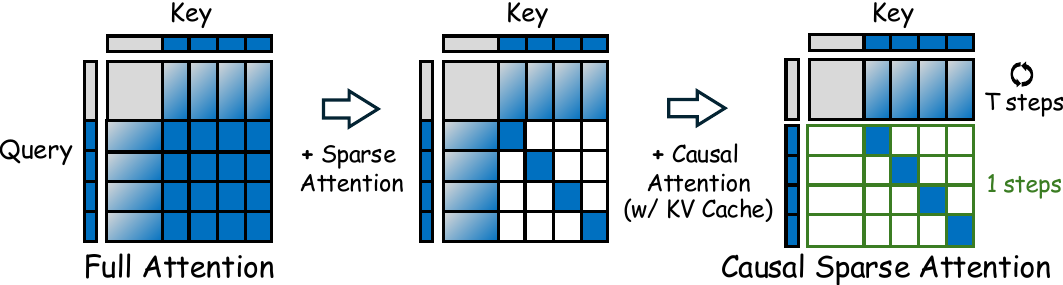}
    % \vspace{-2mm}
    \caption{
    Illustration of the transition from Full Attention to Causal Sparse Attention. This figure highlights the reduction in computational complexity achieved by excluding pairwise calculations among reference images. Additionally, the application of unidirectional causal attention, along with the use of KV-Cache, further enhances computational efficiency while ensuring effective transmission of essential color ID information.
    }
    % \vspace{-2mm}
    \label{fig:attention}
\end{figure}
Prior studies~\cite{zhuang2024colorflow,huang2024context} demonstrate that concatenating images along the spatial dimension improves identity preservation by leveraging the strong contextual matching capabilities of self-attention. However, the high computational overhead of Full Attention limits its practical usage to only a few reference images during both training and inference.

 In our framework, we define the sequence length of the line art latent as \( S_l \) and that of each reference image latent as \( S_r \). The computational complexity of the full attention mechanism across the total time steps \( T \) is given by:
\begin{equation}
\mathcal{O}(T \times (S_l^2 + 2N \times S_l \times S_r + N^2 \times S_r^2)).
\end{equation}
As the number of reference images \( N \) increases, this complexity can become substantial.

However, treating all reference images as complete images during attention calculations is inefficient. Instead, reference images should primarily provide color ID information for the line art, allowing us to eliminate pairwise computations among them.
To mitigate this inefficiency, we transition from a full attention mechanism to a sparse attention mechanism by excluding calculations between reference images, as illustrated in Fig. \ref{fig:attention}. This adjustment reduces computational complexity to:
\begin{equation}
\mathcal{O}(T \times (S_l^2 + 2N \times S_l \times S_r + N \times S_r^2)).
\end{equation}

Furthermore, since the reference images are clean and pre-existing, they do not require a full diffusion denoising process alongside the noise latents. To maintain independence among the reference image latents, we modify the bidirectional attention between the reference images and noise latents to unidirectional causal attention. This also ensures that the reference image latents preserve color information, which is then transferred to the coloring region. Reference image latents only require one denoise step at time step 0. We implement a KV-Cache to store layer-wise keys and values, providing conditional guidance for the noise latents during the diffusion denoising process and ensuring consistent preservation of color IDs, as shown in Fig. \ref{fig:attention}. By transitioning from Sparse Attention to Causal Sparse Attention, we further reduce the computational complexity to:

\begin{equation}
\mathcal{O}(T \times (S_l^2 + N \times S_l \times S_r) + N \times S_r^2).
\end{equation}
In summary, Causal Sparse DiT \( D_{cs} \) enhances the efficiency and effectiveness of the colorization process by leveraging optimized attention mechanisms. This not only reduces computational complexity but also preserves essential color information.

\subsection{Localized Reusable Position Encoding}
\textcolor{black}{Previous works have adopted two main strategies for spatial image concatenation: grid concatenation and horizontal/vertical concatenation. The grid concatenation restricts the number of reference images to specific counts, while horizontal or vertical concatenation often leads to extreme aspect ratios and resolutions, weakening the spatial correspondence between distant reference images and the target region.
Moreover, we observe that the pre-trained PixArt-Alpha model—which supports aspect ratios ranging from 0.25 to 4.0~\cite{chen2023pixart}—struggles to generate coherent results when more than 8 reference images are concatenated horizontally or vertically.}

\textcolor{black}{To address these limitations, we propose Localized Reusable Position Encoding, which enables the integration of an arbitrary number of reference images without modifying existing 2D position encodings. 
As illustrated in Fig.~\ref{fig:FlowChat}, we divide the line art image into four spatial patches (top-left, bottom-left, top-right, and bottom-right) and retrieve the top-$k$ most similar reference images for each patch, resulting in four distinct reference sets.
We patchify the noise latent and all reference latents, obtaining a noise feature $z \in \mathbb{R}^{d,h,w}$ and $N$ reference features $r \in \mathbb{R}^{d,\frac{h}{2},\frac{w}{2}}$.
The positional encoding of shape $(d,h,2w)$ is divided into five parts: one of shape $(d,h,w)$ for the central region, and four of shape $(d,\frac{h}{2},\frac{w}{2})$ for the surrounding regions. The noise feature $z$ is combined with the central positional encoding, while each reference feature $r$ integrates its corresponding local position encoding based on its set.
This is equivalent to concatenating the reference features and the noise feature along the spatial dimension, while reusing the local encodings to maintain their proximity to the central area in positional space. This approach allows for an arbitrary number of references, even exceeding 200.}

During training, we randomly sample images from each reference set, keeping the total number of references fixed at 3, 6, or 12. This enhances the model's adaptability to varying numbers and combinations of reference images.

\subsection{Line Art Guider}
We concatenate the line art images latents and color hint images latents as input to the Line Art Guider \(G \). The features from the Line Art Guider are integrated layer-wise into the main branch, enabling precise control over line art and facilitating flexible incorporation of color hints.

\subsubsection{Self-Attention-Only Block.} Given that the Line Art Guider is solely responsible for receiving image-type control conditions, we eliminate the Cross-Attention layers and retain only the Self-Attention layers. This approach reduces the model's parameters without compromising its control effectiveness.

\subsubsection{Line Art Style Augmentation.}
\begin{figure}[t]
    % \centering
    % \vspace{-3mm}
    \includegraphics[width=0.47\textwidth]{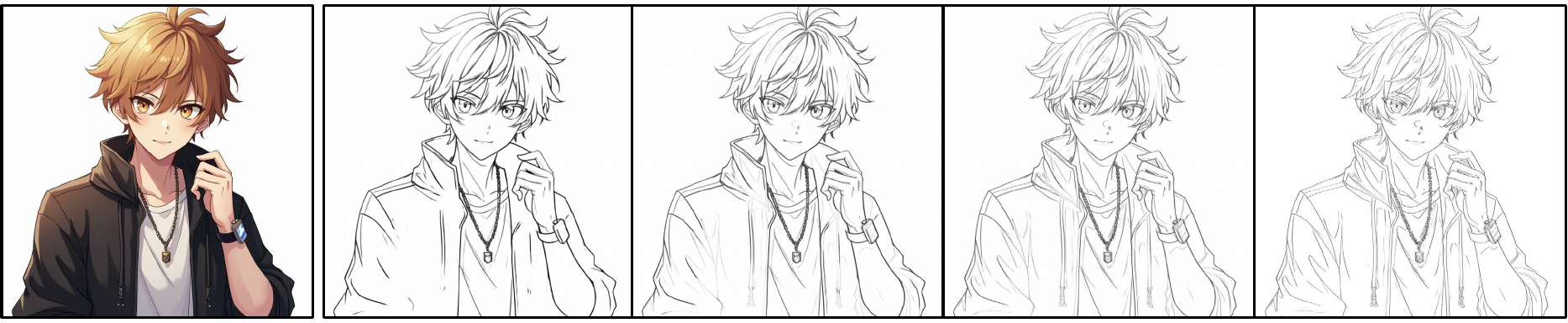}
    \vspace{-2mm}
    \caption{An example of line art style augmentation, demonstrating the blending of outputs from two distinct line art extractors. This strategy improves the robustness of the Line Art Guider to diverse artistic styles.}
    \vspace{-2mm}
    \label{fig:Augmentation}
\end{figure}
We observe that existing line art extraction models produce outputs with distinct styles. To enhance the robustness of the Line Art Guider to varying line art styles, we select two line art extractors\cite{li-2017-deep, Anime2Sketch} with significantly different styles. As shown in Fig. \ref{fig:Augmentation}, during training, we employ line art style augmentation by randomly blending the two styles at varying proportions, thereby improving the Line Art Guider's adaptability to diverse line art styles.

\subsubsection{Hint Point Sampling Strategy.}
\begin{figure}[t]
    % \centering
    % \vspace{-3mm}
    \includegraphics[width=0.38\textwidth]{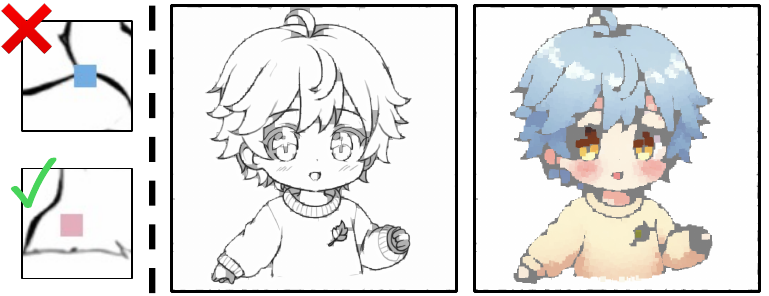}
    \vspace{-2mm}
    \caption{Hint Point Sampling Strategy. This method reduces ambiguity by limiting the RGB pixel value variance within hint points to 0.01, effectively preventing hint points from being placed at edge intersections during training. Additionally, we visualize 30,000 randomly sampled hint points to demonstrate their distribution.}
    \vspace{-3mm}
    \label{fig:hint_point}
\end{figure}
Color hint allows users to specify colors for specific regions directly. To avoid ambiguities during training, such as those illustrated in Fig. \ref{fig:hint_point} where hint points fall on edge intersections, we propose a simple hint points sampling strategy. By constraining the variance of RGB pixels value within the hint points to no more than 0.01, we effectively exclude hint points sampled on edges during the training phase.

The optimization objective for Cobra can be expressed as follows:
\begin{equation}
\mathcal{L} = E_{t, \epsilon} \| \epsilon - D_{cs}(G(Z_L, Z_C, M, t), Z_R^{0:N-1}, Z_t, t) \|^2_2.
\end{equation}
Here, \( \epsilon \) represents Gaussian noise, \( Z_L \) is the latent of the line art image, \( Z_C \) is the latent of the hint color image, \( M \) is the hint mask, \( Z_R^{0:N-1} \) denotes the latents of \( N \) reference images, and \( Z_t \) is the latent of the original image after undergoing the forward diffusion process at the \( t \)-th timestep. Additionally, we adopt a Guided Super-Resolution Pipeline (GSRP), following the methodology of ColorFlow~\cite{zhuang2024colorflow}, to rectify structural distortions caused by the imprecise reconstruction of the Variational Autoencoder.

\section{Experiments}\label{sec:experiments}
\subsection{Dataset and Benchmark}

% We trained our model using a large-scale dataset of over 1.2 million color images from more than 50,000 publicly available comic chapter sequences sourced from various online repositories. After filtering and cleaning, we used the CLIP image encoder \cite{radford2021learning} to annotate at least 12 relevant reference images per comic page based on image similarity, minimizing redundant computations during training.
\subsubsection{Training Dataset.} 
We trained our model using a large-scale dataset of over 1.2 million color images from more than 50,000 comic chapter sequences sourced from various online repositories, including \footnote{\url{https://digitalcomicmuseum.com}} and \footnote{\url{https://comicbookplus.com}}, all in the public domain. After filtering and cleaning, we used the CLIP image encoder \cite{radford2021learning} to annotate at least 12 relevant reference images per comic page based on image similarity, minimizing redundant computations during training.
\subsubsection{Cobra-Bench.} To evaluate \ModelName comprehensively, we established Cobra-Bench, a benchmark of 30 comic chapters excluded from training. Each chapter includes 100 reference images and 50 line art pages, available in standard and shadowed forms to mimic traditional black-and-white comics. 
We used five metrics for quantitative assessment against baseline methods: three perceptual measures—CLIP Image Similarity (CLIP-IS)\cite{radford2021learning}, Fréchet Inception Distance (FID)\cite{fid}, and Aesthetic Score (AS)\cite{schuhmann2022laion}—and two pixel-wise metrics—Peak Signal-to-Noise Ratio (PSNR)\cite{psnr} and Structural Similarity Index (SSIM)~\cite{ssim}. These metrics collectively provide a robust evaluation of colorization quality, assessing both aesthetic appeal and fidelity to the original content.

\subsection{Implementation Details}
\ModelName~ is trained based on the pretrained PixArt-Alpha~\cite{chen2023pixart}. The trainable components of \ModelName~ include the Line Art Guider and a LoRA applied to the pretrained Full Attention DiT, which facilitates the transition to our proposed Causal Sparse DiT. We conducted a total of 78,000 training steps with a learning rate of 1e-5 and a batch size of 16, utilizing the AdamW optimizer. The training resolution was set to $640\times1024$ (width $\times$ height), while the resolution of the reference images was 320$\times$512 (width $\times$ height). Additionally, since \ModelName~ does not rely on text conditions, we utilized the empty text output from the text encoder as input during both the training and inference phases.

\subsection{Comparison with baselines}
\textcolor{black}{In this section, we conduct a quantitative comparative analysis of \ModelName~ against leading state-of-the-art comic colorization methods. Specifically, we evaluate ColorFlow~\cite{zhuang2024colorflow}, a diffusion-based colorization model that leverages reference images, and MangaColorization v2 (MC-v2)~\cite{mcv2}, a GAN-based model that operates without references. Additionally, we include comparisons with a generative approache that offer similar functionalities. 
Specifically,we utilize a pre-trained IP-Adapter~\cite{ye2023ip} to convert reference images into image prompts, combined with a pre-trained ControlNet~\cite{zhang2023adding} that guides Stable Diffusion 1.5~\cite{ldm22} for the comic line art colorization.
In our experiments, We fixed the number of reference images for \ModelName~ at 24. Both ColorFlow and \ModelName~ were configured with an inference step of 10, while the IP-Adapter was set to the official default of 50 steps. To ensure a fair comparison, all images were resized to a resolution of 512×800, consistent with ColorFlow's default inference resolution, prior to evaluating the PSNR and SSIM.}

\subsubsection{Qualitative results.}
\begin{figure*}[t]
    % \centering
    % \vspace{-3mm}
    \includegraphics[width=0.8\textwidth]{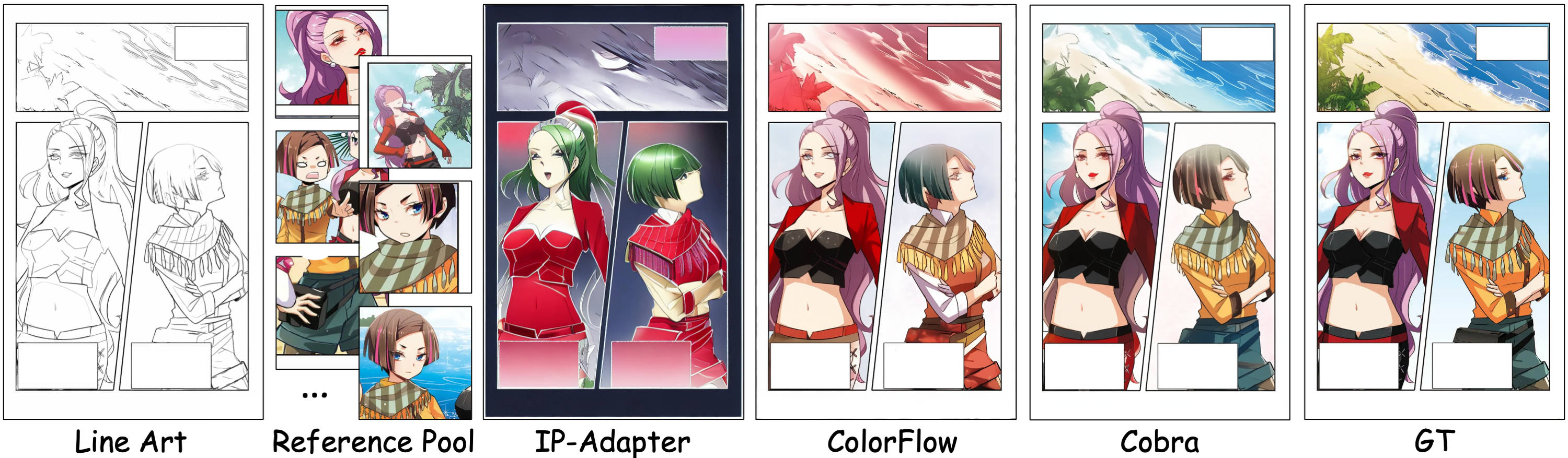}
    % \vspace{-2mm}
    \caption{Qualitative results of line art colorization, highlighting how \ModelName~ outperforms other methods by accurately preserving color IDs and providing high-quality results, even in complex scenarios.}
    \vspace{-2mm}
    \label{fig:exp_line}
\end{figure*}
\begin{figure*}[t]
    % \centering
    % \vspace{-3mm}
    \includegraphics[width=0.8\textwidth]{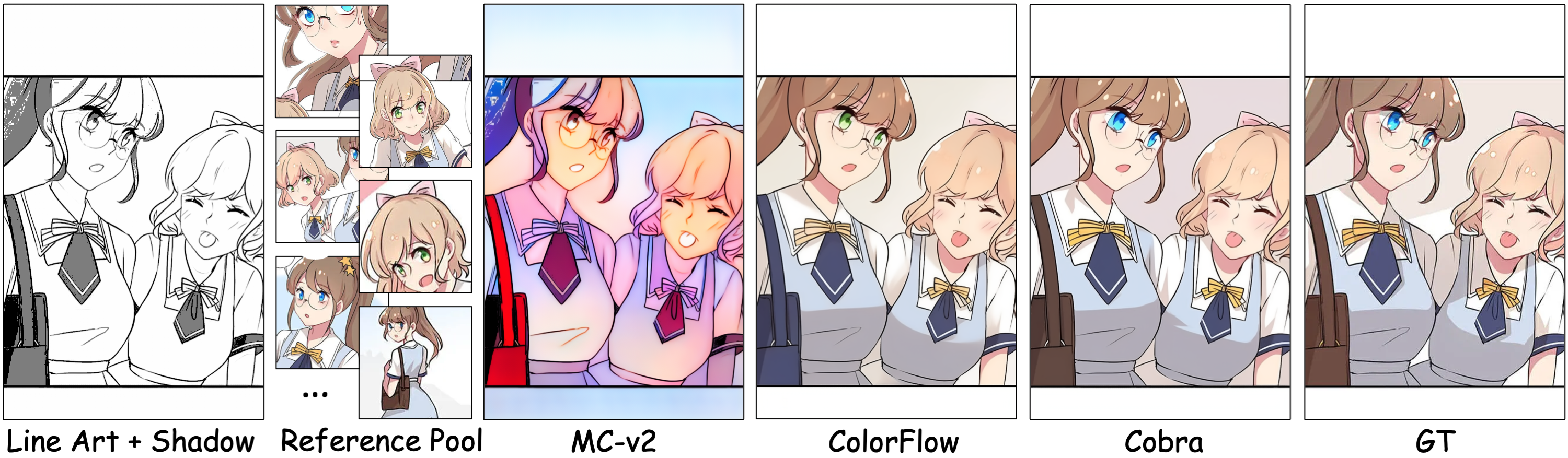}
    % \vspace{-2mm}
    \caption{Qualitative results of line art with shadows, showcasing \ModelName~'s superior ability to maintain color fidelity and enhance detail, demonstrating its robustness and effectiveness in real-world applications.}
    \vspace{-2mm}
    \label{fig:exp_bwg}
\end{figure*}
\begin{figure*}[t]
    % \centering
    % \vspace{-3mm}
    \includegraphics[width=0.95\textwidth]{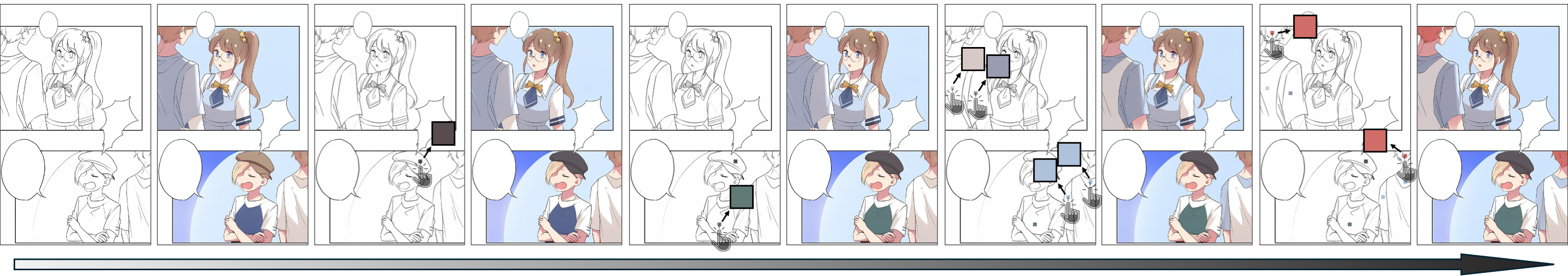}
    % \vspace{-2mm}
    \caption{Interactive line art colorization using color hints, demonstrating how \ModelName~ allows users to specify color adjustments in designated areas, enhancing control over the colorization process while maintaining overall stability.}
    % \vspace{-2mm}
    \label{fig:hint}
\end{figure*}
As illustrated in Fig. \ref{fig:exp_line} and \ref{fig:exp_bwg}, the approach employed by the IP-Adapter, which maps reference images to coarse image prompts, results in the blending of color IDs, rendering it incapable of achieving reasonable colorization for line art images. Although MC-v2 is capable of comic colorization, its lack of reference information leads to inaccuracies in character coloring, alongside issues of excessive color saturation.
Additionally, while ColorFlow can generally provide accurate colorization for line art, its limited number of reference images often results in the omission of critical color IDs, leading to erroneous colors for certain objects or characters. 
Conversely, \ModelName~ effectively references a richer set of contextual images, enabling robust and comprehensive extraction of color IDs. This capability facilitates a more refined and higher-quality result in comic line art colorization, demonstrating its superiority in the practical industrial applications.

\subsubsection{Quantitative results. }
\begin{table*}[htbp]
    \centering
  \caption{Quantitative comparison of \ModelName~ with state-of-the-art comic colorization methods, including CLIP-IS~\cite{radford2021learning}, FID~\cite{fid}, PSNR~\cite{psnr}, SSIM~\cite{ssim}, and AS~\cite{schuhmann2022laion} metrics for both line art and line art with shadows.}
  \vspace{-2mm}
  \label{tab:exp1}
\resizebox{0.88\textwidth}{!}{
      \begin{tabular}{c|c|c c c c c|c c c c c}
    \toprule[1.5pt]
  \multirow{2}{*}{Method}  & \multirow{2}{*}{Reference-based} & \multicolumn{5}{c|}{Line Art}   & \multicolumn{5}{c}{Line Art + Shadow}\\
     &   & CLIP-IS$\uparrow$ & FID$\downarrow$  & PSNR$\uparrow$ &SSIM$\uparrow$ &AS$\uparrow$ &  CLIP-IS$\uparrow$ & FID$\downarrow$  & PSNR$\uparrow$ &SSIM$\uparrow$ &AS$\uparrow$\\ \toprule
     MC-v2 & &  - & - & - &- & - &  0.8635 & 44.09 & 15.57 &0.7630 & 4.531\\ \midrule
    IP-Adapter & \checkmark&  0.8284 & 76.01 & 8.111 &0.5561 & 4.511 &  - & - & - &- & -\\ \midrule
    ColorFlow & \checkmark &  0.9030 & 26.29 & 15.20 &0.8045 & 4.630 &  0.9198 & 21.79 & 18.49 &0.8631 & 4.657\\ \midrule
    Cobra  & \checkmark & \textbf{0.9176}  & \textbf{20.98} & \textbf{16.08} &\textbf{0.8141} & \textbf{4.641} &  \textbf{0.9264} & \textbf{18.84} & \textbf{18.96} &\textbf{0.8694} & \textbf{4.674}\\
    \bottomrule[1.5pt]
  \end{tabular}
}
\vspace{-2mm}
\end{table*}
Tab. \ref{tab:exp1} summarizes the quantitative results. The IP-Adapter performs poorly in line art colorization, significantly trailing both ColorFlow and \ModelName~ across all five evaluation metrics. In contrast, \ModelName~ excels in identity preservation and image quality by utilizing richer contextual information. For line art with shadow colorization, the non-reference MC-v2 consistently underperforms compared to reference-based methods in both perceptual and pixel-wise metrics. Notably, \ModelName~ achieves the best performance across all evaluation criteria, underscoring its robustness in comic colorization. Additional results pertaining to line art colorization for an anime video are provided in the Appendix.

\subsection{Contributions over ColorFlow}
\textcolor{black}{\ModelName offers several improvements over ColorFlow. It supports arbitrary numbers of references via Localized Reusable Position Encoding, whereas ColorFlow is limited to 12 due to grid concatenation. \ModelName also simplifies the architecture by directly concatenating multiple references with the noise latent and processing them through a Causal Sparse DiT, improving both efficiency and image quality. Additionally, \ModelName supports color hints for enhanced user control.}

\textcolor{black}{As shown in Tab.~\ref{tab:colorflow_comparison}, \ModelName outperforms ColorFlow across all metrics with 12 references at 640×1024 resolution, while requiring significantly less time and memory. Even with only 2 references \ModelName still achieves superior results.}

\begin{table}[h]
\centering
\caption{Quantitative comparison of ColorFlow and Cobra on colorization.}
\label{tab:colorflow_comparison}
\vspace{-2mm}
\resizebox{0.5\textwidth}{!}{
\begin{tabular}{l|ccccc|cc}
\toprule[1.5pt]
Method & CLIP-IS & FID$\downarrow$ & PSNR$\uparrow$ & SSIM$\uparrow$ & AS$\uparrow$ & Time(s) & Mem(GB) \\
\midrule
ColorFlow (12 refs) & 0.9030 & 26.29 & 15.20 & 0.8045 & 4.630 & 1.03 & 36.4 \\
Cobra (12 refs)     & \textbf{0.9132} & \textbf{21.86} & \textbf{15.94} & \textbf{0.8136} & \textbf{4.642} & \textbf{0.31} & \textbf{9.3} \\
\midrule
ColorFlow (2 refs)  & 0.8798 & 32.55 & 14.88 & 0.8013 & 4.591 & - & - \\
Cobra (2 refs)      & \textbf{0.8989} & \textbf{26.91} & \textbf{15.19} & \textbf{0.8120} & \textbf{4.611} & - & - \\
\bottomrule[1.5pt]
\end{tabular}
}
\vspace{-2mm}
\end{table}

\subsection{Colorization Using Color Hints}
Fig.~\ref{fig:hint} demonstrates interactive line art colorization using user-provided color hint points. When a large set of reference images doesn't fully meet user needs, \ModelName~ allows users to enhance colorization by adding color hints in specific areas. As shown in the interaction, when a user specifies a color hint, \ModelName~ accurately adjusts the color in that region, demonstrating fidelity to user input and offering precise control for practical applications. The model maintains stability and consistency in areas not influenced by hints.
\subsection{User Study}
\begin{table}[htbp]
    \centering
  \caption{
  Results of the User Study. The table presents the voting rates for \ModelName~and ColorFlow based on contextual color IDs consistency, plausibility of object colors, and overall aesthetic quality.
  }
  \vspace{-2mm}
  \label{tab:user}
\resizebox{0.5\textwidth}{!}{
      \begin{tabular}{c | c c c}
    \toprule[1.5pt]
      &  Color IDs consistency$\uparrow$  & Color plausibility$\uparrow$ &Aesthetic quality$\uparrow$\\ \toprule
     \ModelName~ & \textbf{79.1\%} & \textbf{69.3\%} & \textbf{73.2\%} \\ \midrule
    ColorFlow   &  20.9\% & 30.7\% & 26.8\%\\
    \bottomrule[1.5pt]
  \end{tabular}
}
\end{table}
In the user study, we conducted a comparative analysis of \ModelName~ and the suboptimal model ColorFlow, as evaluated in quantitative experiments, across three dimensions: contextual color IDs consistency, plausibility of object colors, and overall aesthetic quality. We collected over 4,000 valid votes, with the results presented in Tab. \ref{tab:user}. These findings demonstrate a clear preference for \ModelName~ in all assessed aspects and provide robust evidence of its superior performance in comic colorization.
\subsection{Ablation Study}
In this section, we will analyze the impact of the number of reference images on the quality of line art colorization and the effectiveness of Causal Sparse Attention in accelerating inference. 

\subsubsection{Impact of Reference Image Count.}
\begin{table}[t]
    \centering
    \vspace{1mm}
  \caption{Impact of reference image count on the performance of \ModelName~ in line art colorization, demonstrating consistent improvements as the number of reference images increases.}
  \vspace{-2mm}
  \label{tab:ref_num}
\resizebox{0.4\textwidth}{!}{
      \begin{tabular}{c | c c c c c}
    \toprule[1.5pt]
     Number &  CLIP-IS$\uparrow$ & FID$\downarrow$  & PSNR$\uparrow$ &SSIM$\uparrow$ &AS$\uparrow$\\ \toprule
    4 &  0.9083  &  23.18 & 15.61  & 0.8133 & 4.634 \\ \midrule
    12  &  0.9132 & 21.86 & 15.94 & 0.8136 & 4.642 \\ \midrule
    24  &  0.9176 & 20.98 & 16.08 & 0.8141 & 4.641 \\ \midrule
    36  &  \textbf{0.9183} & \textbf{20.64} & \textbf{16.13} & \textbf{0.8142} & \textbf{4.649} \\ 
    \bottomrule[1.5pt]
  \end{tabular}
}
\end{table}
In this study, we investigate how the number of reference images impacts line art colorization quality using the Cobra-Bench evaluation framework. We evaluated \ModelName~ with 4, 12, 24, and 36 reference images, as shown in Tab.~\ref{tab:ref_num}. \textcolor{black}{Our results show that as the number of reference images increases, colorization accuracy improves, with better preservation of small but important details like character accessories and eye color. These details, though occupying small areas, are critical for high-demand applications.
Tab.~\ref{tab:ref_num} illustrates a steady improvement in all metrics as the number of reference images increases, highlighting the importance of having a larger and more diverse set of reference images to optimize comic line art colorization.}

\subsubsection{Effectiveness of Causal Sparse Attention.}
\begin{table}[t]
    \centering
    \vspace{1mm}
  \caption{Performance comparison of three attention variants (Full, Sparse, and Causal Sparse Attention) with 24 reference images under FP16 precision.}
  \vspace{-2mm}
  \label{tab:efficiency}
\resizebox{0.48\textwidth}{!}{
      \begin{tabular}{c | c c c}
    \toprule[1.5pt]
     Attention Type & Time(s)$/$step$\downarrow$ & Computation(TFlops)$\downarrow$ &	FID$\downarrow$\\ \toprule
    Full Attention & 1.99 & 38.2 & -  \\ \midrule
    Sparse Attention & 0.81 & 14.7 & 21.07   \\ \midrule
    Causal Sparse Attention  & \textbf{0.35} & \textbf{9.0} & \textbf{20.98}  \\
    \bottomrule[1.5pt]
  \end{tabular}}

\end{table}
\begin{figure}[t]
    % \centering
    % \vspace{-3mm}
    \includegraphics[width=0.38\textwidth]{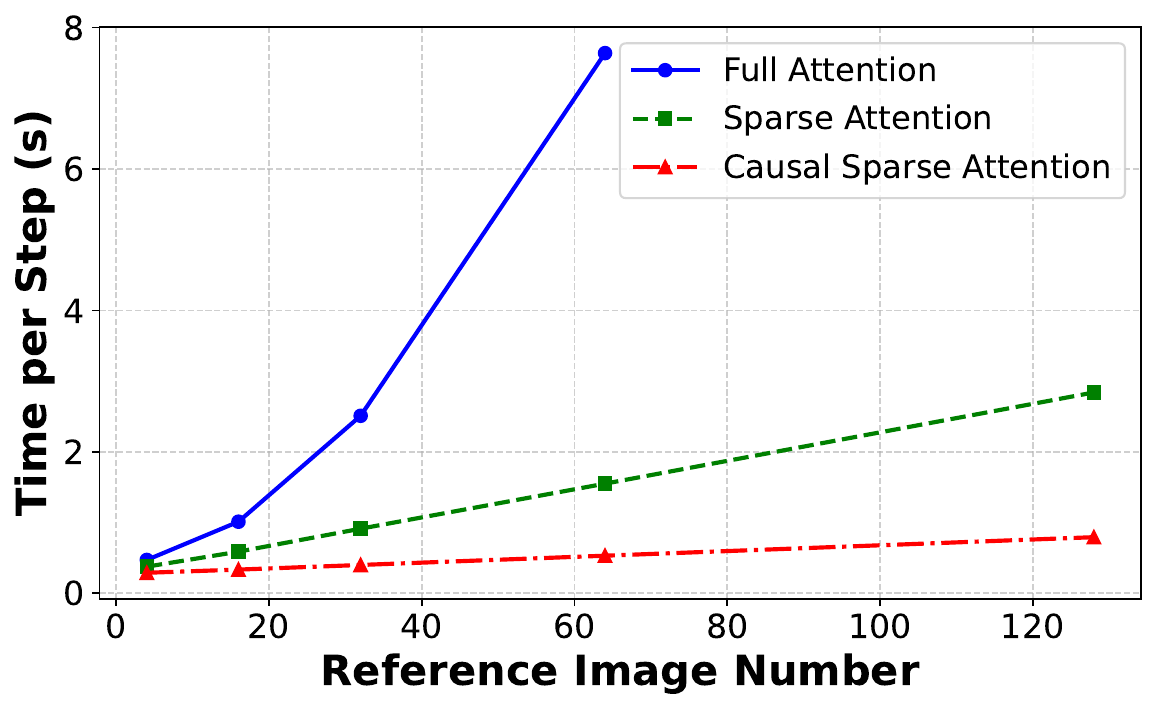}
    \vspace{-4mm}
    \caption{Evaluation of inference time efficiency for Full Attention, Sparse Attention, and Causal Sparse Attention (with KV-Cache) across different counts of reference images.}
    \vspace{-2mm}
    \label{fig:time}
\end{figure}
\textcolor{black}{We evaluate three attention mechanisms—Full Attention, Sparse Attention, and Causal Sparse Attention (with KV-Cache)—under a fixed setting of 24 reference images and FP16 precision. As reported in Tab.~\ref{tab:efficiency}, Causal Sparse Attention achieves significantly better inference efficiency than the other two, while maintaining comparable colorization quality.}
To assess scalability, we further measure inference latency with different numbers of reference images: 4, 16, 32, 64, and 128 (Fig.~\ref{fig:time}). Full Attention exhibits quadratic growth in inference time, becoming approximately 15× slower than Causal Sparse Attention at 64 references.
Causal Sparse Attention also consistently outperforms Sparse Attention, requiring only one-third of its latency at 64 references.
These results demonstrate the substantial efficiency of Causal Sparse Attention, making it a more favorable choice for practical line art colorization applications.

\subsubsection{Limitation.}
\begin{figure}[t]
    % \centering
    % \vspace{-3mm}
    \includegraphics[width=0.45\textwidth]{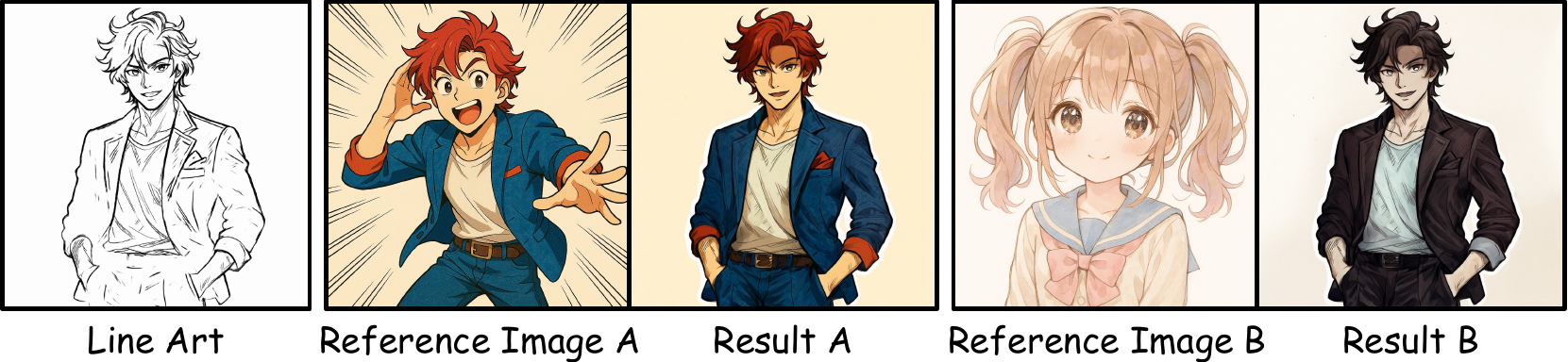}
    \vspace{-4mm}
    \caption{Limitation of \ModelName~. \ModelName~ preserves color identity when references share the same character (Result A), but fails when references depict different characters (Result B).}
    \vspace{-2mm}
    \label{fig:limitation}
\end{figure}
\textcolor{black}{
As shown in Fig.~\ref{fig:limitation}, \ModelName~ can handle slight variations in art style when transferring colors from references of the same character. However, the model is explicitly designed to transfer consistent color identities for the same character and fails to generalize when references depict different characters, limiting its ability to perform style transfer across identities.}

\section{Conclusion}\label{sec:conclusion}
In this work, we introduce \ModelName, a novel long-context framework for comic line art colorization. By leveraging a rich set of reference images, our approach enhances colorization accuracy while preserving fine-grained identity details. With components like Causal Sparse DiT and Localized Reusable Position Encoding, \ModelName significantly improves efficiency and color fidelity over existing methods.
Extensive qualitative and quantitative evaluations demonstrate its superior performance, especially in industrial applications where user control and contextual richness are essential. Overall, \ModelName represents a major advancement in comic colorization, providing robust solutions for diverse artistic styles and user needs.

\begin{acks}
This work was supported by the National Key R\&D Program of China (2022YFB4701400/4701402), SSTIC Grant(KJZD20230923115106012, KJZD20230923114916032, GJHZ20240218113604008) and Beijing Key Lab of Networked Multimedia.
\end{acks}

%%
%% The next two lines define the bibliography style to be used, and
%% the bibliography file.
\bibliographystyle{ACM-Reference-Format}
\bibliography{sample-base}

\begin{figure*}[t]
    % \centering
    % \vspace{-3mm}
    \includegraphics[width=1.0\textwidth]{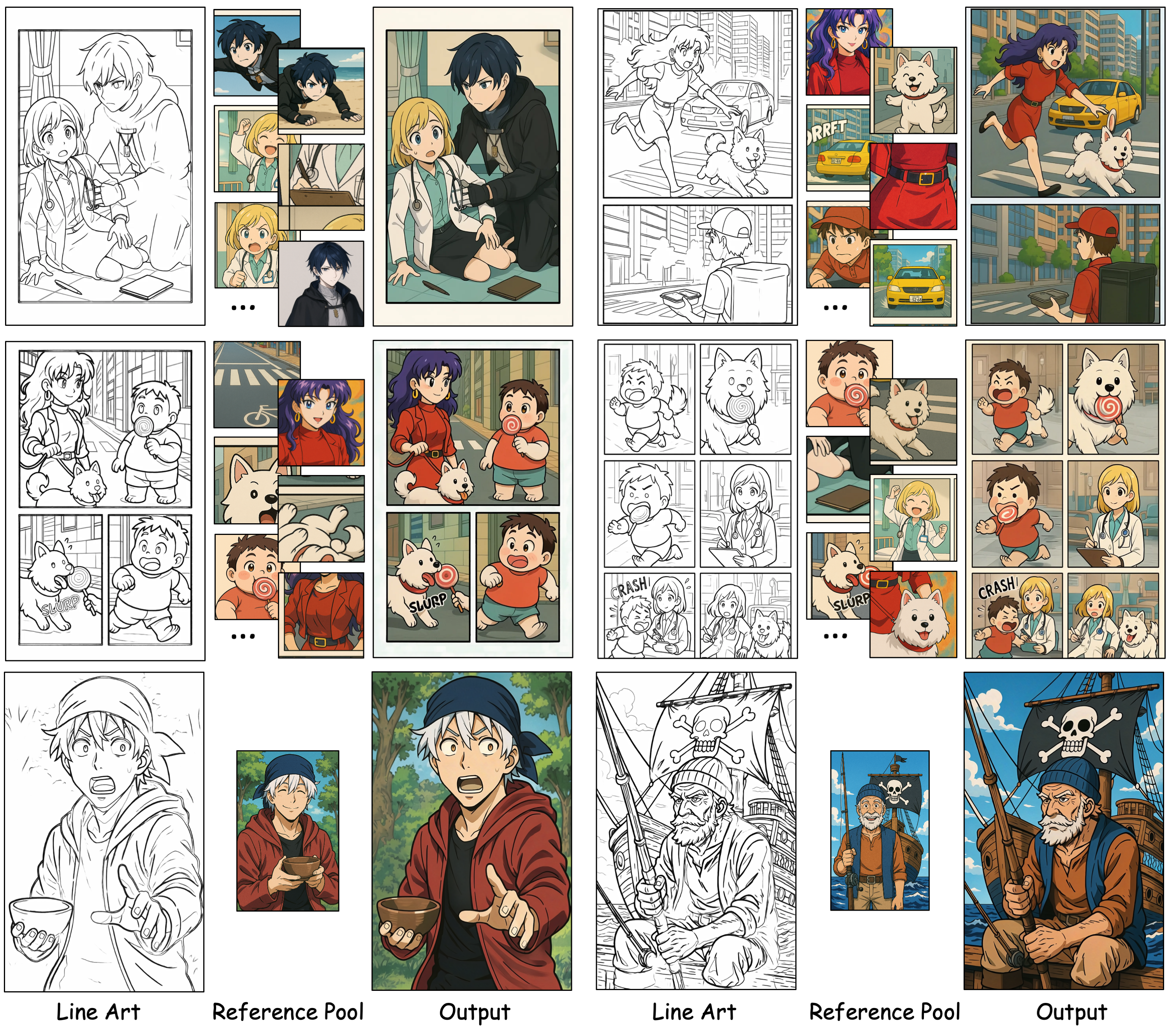}
    % \vspace{-2mm}
    \caption{Additional results of colorization for comic line art.}
    % \vspace{-2mm}
    \label{fig:supp_line}
\end{figure*}
\begin{figure*}[t]
    % \centering
    % \vspace{-3mm}
    \includegraphics[width=1\textwidth]{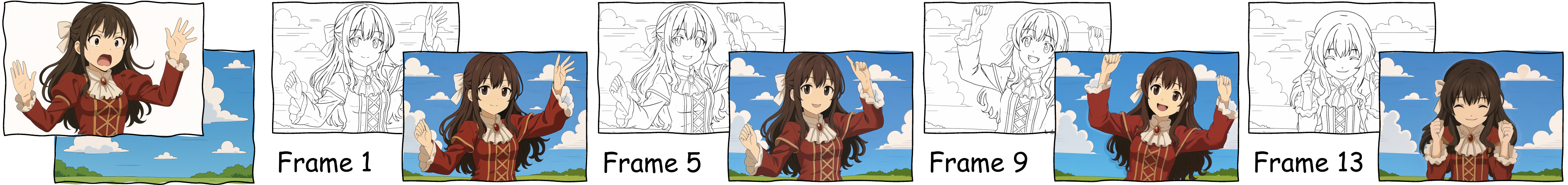}
    % \vspace{-2mm}
    \caption{Additional results of line art colorization for an anime video.}
    % \vspace{-2mm}
    \label{fig:supp_line}
\end{figure*}
\begin{figure*}[t]
    % \centering
    \vspace{2mm}
    \includegraphics[width=1.0\textwidth]{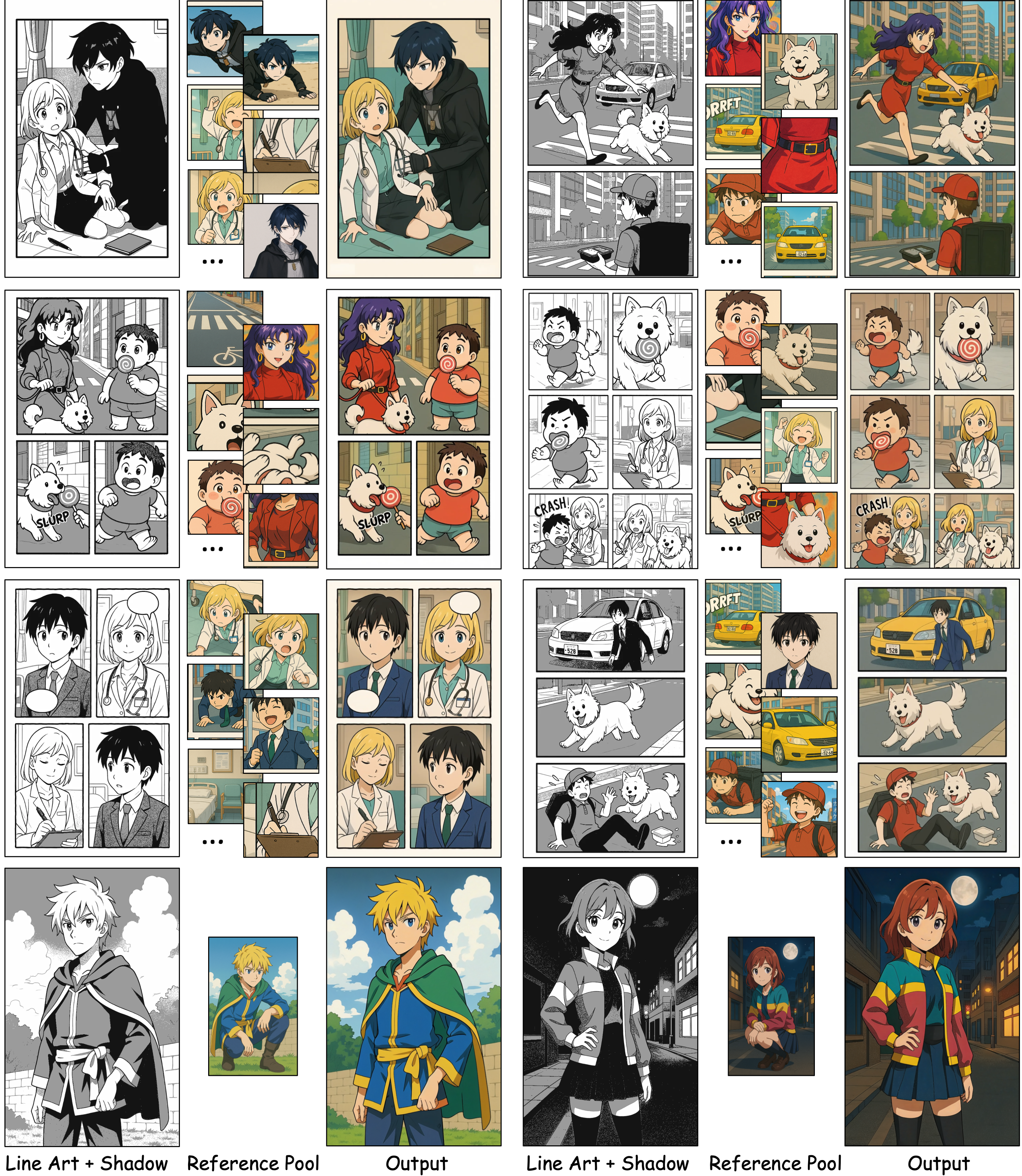}
    % \vspace{-2mm}
    \caption{Additional results of colorization for comic line art with shadow.}
    % \vspace{-2mm}
    \label{fig:supp_line}
\end{figure*}
%%
%% If your work has an appendix, this is the place to put it.
\appendix

% \section{Research Methods}

% \subsection{Part One}

% Lorem ipsum dolor sit amet, consectetur adipiscing elit. Morbi
% malesuada, quam in pulvinar varius, metus nunc fermentum urna, id
% sollicitudin purus odio sit amet enim. Aliquam ullamcorper eu ipsum
% vel mollis. Curabitur quis dictum nisl. Phasellus vel semper risus, et
% lacinia dolor. Integer ultricies commodo sem nec semper.

% \subsection{Part Two}

% Etiam commodo feugiat nisl pulvinar pellentesque. Etiam auctor sodales
% ligula, non varius nibh pulvinar semper. Suspendisse nec lectus non
% ipsum convallis congue hendrerit vitae sapien. Donec at laoreet
% eros. Vivamus non purus placerat, scelerisque diam eu, cursus
% ante. Etiam aliquam tortor auctor efficitur mattis.

% \section{Online Resources}

% Nam id fermentum dui. Suspendisse sagittis tortor a nulla mollis, in
% pulvinar ex pretium. Sed interdum orci quis metus euismod, et sagittis
% enim maximus. Vestibulum gravida massa ut felis suscipit
% congue. Quisque mattis elit a risus ultrices commodo venenatis eget
% dui. Etiam sagittis eleifend elementum.

% Nam interdum magna at lectus dignissim, ac dignissim lorem
% rhoncus. Maecenas eu arcu ac neque placerat aliquam. Nunc pulvinar
% massa et mattis lacinia.

\end{document}